\documentclass[journal]{IEEEtran}

\IEEEoverridecommandlockouts                              

\usepackage{times}

\usepackage{soul}
\usepackage{url}
\usepackage[hidelinks]{hyperref}
\usepackage[small]{caption}
\usepackage{booktabs}
\usepackage{amsfonts}
\usepackage{enumitem}

\usepackage{hyperref}
\hypersetup{colorlinks, linkcolor ={blue}, citecolor={green}, urlcolor={red}}
\usepackage{cite}
\usepackage{amsmath,amssymb,amsfonts}
\usepackage{algorithmic}
\usepackage{graphicx}
\usepackage{textcomp}
\usepackage{xcolor}
\def\BibTeX{{\rm B\kern-.05em{\sc i\kern-.025em b}\kern-.08em
    T\kern-.1667em\lower.7ex\hbox{E}\kern-.125emX}}

\title{\Large \bf
Interaction-aware Kalman Neural Networks for Trajectory Prediction
}

\author{Ce Ju$^{{1, \dagger}}$, Zheng Wang$^{2,\dagger}$, Cheng Long$^2$, Xiaoyu Zhang$^{3,4}$ and Dong Eui Chang$^5$
\thanks{$\dagger$ Equal Contribution}
\thanks{$^{1}$WeBank Co., Ltd., AI Department, {\tt\small ceju@webank.com}}
\thanks{$^{2}$Nanyang Technological University, Computer Science and Engineering Department, {\tt\small wang\_zheng@ntu.edu.sg, c.long@ntu.edu.sg}}
\thanks{$^{3}$University of Michigan, Ann Arbor, Electrical and Computer Engineering Department, {\tt\small zhxiaoyu@umich.edu}}
\thanks{$^{4}$Shenzhen Institute of Artificial Intelligence and Robotics for Society}
\thanks{$^{5}$Korea Advanced Institute of Science and Technology, Electrical Engineering Department, {\tt\small dechang@kaist.ac.kr}}
\thanks{*This research was in part supported by WeBank Co., Ltd., the Institute for Information \& Communications Technology Planning \& Evaluation(IITP) grant funded by the Korea government(MSIT) (No. 2019-0-01396, Development of framework for analyzing, detecting, mitigating of bias in AI model and training data), and by the ICT R\&D program of MSIP/IITP [2016-0-00563, Research on Adaptive Machine Learning Technology Development for Intelligent Autonomous Digital Companion].}
}

\begin{document}

\maketitle

\begin{abstract}
Forecasting the motion of surrounding obstacles (vehicles, bicycles, pedestrians and etc.) benefits the on-road motion planning for intelligent and autonomous vehicles. 
Complex scenes always yield great challenges in modeling the patterns of surrounding traffic. 
For example, one main challenge comes from the intractable interaction effects in a complex traffic system. 
In this paper, we propose a multi-layer architecture \emph{Interaction-aware Kalman Neural Networks} (IaKNN) which involves an interaction layer for resolving high-dimensional traffic environmental observations as interaction-aware accelerations, a motion layer for transforming the accelerations to interaction-aware trajectories, and a filter layer for estimating future trajectories with a Kalman filter network. 
Attributed to the multiple traffic data sources, our end-to-end trainable approach technically fuses dynamic and interaction-aware trajectories boosting the prediction performance.
Experiments on the NGSIM dataset demonstrate that IaKNN outperforms the state-of-the-art methods in terms of effectiveness for traffic trajectory prediction.
\end{abstract}

\section{Introduction}

Autonomous driving systems can be broadly categorized into three hierarchical subsystems, namely \emph{perception/localization}, \emph{planning} and \emph{control}~\cite{urmson2008autonomous}. 
The perception subsystem refers to the ability to acquire information from the environment via multiple vehicle sensors like GPS, LiDAR, RADAR, and Camera. It categorizes sensor data by their semantic meaning.
The localization subsystem determines the global and local position of ego-vehicle with respect to High Definition Map and the vehicle's coordinate system respectively.
The planning subsystem typically includes mission planning, behavioral planning, and motion planning, generating an efficient and safe trajectory (specific positions and associated target velocities) for the autonomous vehicle to follow from a start waypoint to a goal waypoint~\cite{wei2014behavioral}.
The control subsystem refers to the ability to execute the planned actions in trajectories from planning by sending accelerations, brake, and steering messages to the actuator on intelligent vehicles. 

Each subsystem in an autonomous driving system has technical difficulties from both hardware and software. 
For example, one hardcore problem for the planning is of generating ego-vehicle trajectories based on low-resolution perception information captured and estimated by the low-cost sensing subsystem. 
Moreover, forecasting the motion of surrounding static and dynamic obstacles is also a big challenge to most of the on-road autonomous driving systems. 
Attributed to the tremendous progress in recent years, the planning with static obstacles has been adequately solved~\cite{mcnaughton2011parallel,mcnaughton2011motion,xu2012real}, while the planning with dynamic obstacles is still explored. 
%
%
A practical and efficient solution to this challenge, which is widely adopted in the autonomous industry such as Baidu's open-source software platform for autonomous vehicles development (Apollo)~\cite{fan2018baidu}, is technically to make the on-road dynamic obstacles become almost static. 

\begin{figure}
	\centering
	\includegraphics[width=0.4\textwidth]{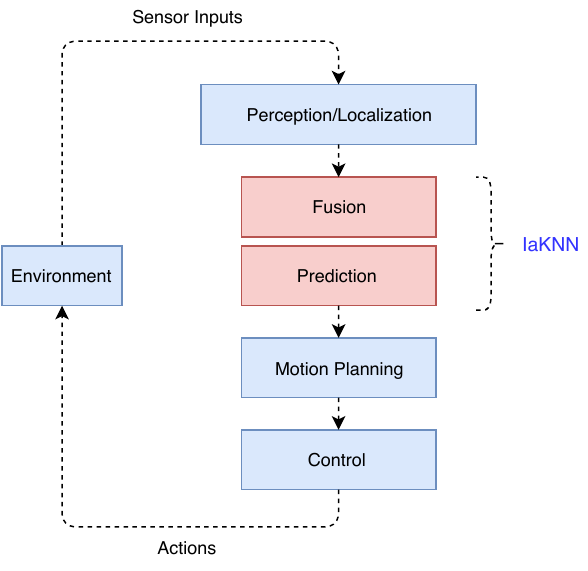}
	\caption{A Illustration of autonomous driving system overview}
	\vspace{-2mm}
	\label{system}
\end{figure}
To handle the above challenges, an independent subsystem \emph{prediction} is laid after \emph{perception} and before \emph{motion planning} to weaken the risk of the on-road planing with dynamic obstacles by predicting the future motion of obstacles, referring to Figure \ref{system}.
Specifically, an effective prediction subsystem needs to handle the on-road challenges including noisy sensing information and complex traffic scenes.
Existing on-road prediction subsystem is categorized into three models, namely the physics-based motion model, the maneuver-based motion model, and the interaction-aware motion model~\cite{liu2005learning,frazzoli2005maneuver,lefevre2014survey}.
The physics-based motion model is the one based on the kinematics.
The maneuver-based motion model is the one designed for a particular maneuver in which the future trajectory of a vehicle is predicted by searching the trajectories which have been clustered as a priori.
The interaction-aware motion model is the one which captures the interactive effects among vehicles by predicting the trajectories of multiple vehicles collectively. Many up-to-date interaction-aware motion models adopt deep learning approach~\cite{alahi2016social,gupta2018social,deo2018convolutional,kuefler2017imitating,bhattacharyya2018multi,ma2018trafficpredict}.

In this paper, we propose a specific model for the prediction subsystem called \emph{Interaction-aware Kalman Neural Networks} (IaKNN).
IaKNN is a multi-layer architecture consisting of three layers, namely an interaction layer, a motion layer, and a filter layer.
The interaction layer is a deep neural network with multiple convolution layers laying before the LSTM encoder-decoder architecture.
Fed with the past trajectories of vehicles that are close to one another, this layer extracts the \emph{accelerations} that capture not only those raw acceleration readings but also the interactive effects among vehicles in the form of social force, a latent variable (which is a measure of internal motivation of an individual in a social activity in sociology and has been used for studying the motion trajectories of pedestrians \cite{helbing1995social}).
The extracted accelerations are called \emph{interaction-aware accelerations}. The motion layer is similar to the existing physics-based motion model which transforms accelerations into trajectories by using kinematics models.
Here, instead of feeding the motion layer with the accelerations read from sensors directly, we feed with those interaction-aware accelerations that are outputted by the interaction layer and call the transformed trajectories \emph{interaction-aware trajectories}.
The filter layer consists of mainly a Kalman filter for optimally estimating the future trajectories based on the interaction-aware trajectories outputted by the motion layer.
The novelty in this layer is that we incorporate two LSTM neural networks~\cite{hochreiter1997long} for learning the time-varying process and measurement noises that would be used in the update step of the Kalman filter, and this is the first of its kind for trajectory prediction.
In summary, our IaKNN model enjoys the merits of both the physics-based model (the motion layer) and the interaction-based model (the interaction layer) and employs neural-network-based probabilistic filtering for accurate estimation (the filter layer). In experiments, we evaluate IaKNN on the \emph{Next Generation Simulation} (NGSIM) dataset \cite{colyar2007us} and the empirical results demonstrate the effectiveness of IaKNN.

In summary, the major contributions of this paper are listed as follows: our approach, to the best of our knowledge, is the first neural network-based filtering algorithm for on-road trajectory prediction, which is end-to-end trainable to learn the time-varying process and measurement noises with LSTM neural networks in a Kalman filter. 
For the normal framework of an on-road autonomous driving system, the value of our methodology is the integration of the techniques in sensor fusion and data-driven approach motion prediction, referring to Figure \ref{system}, more practical to the problem in a dynamic system.
We perform extensive experiments on the NGSIM dataset, which shows that IaKNN consistently outperforms the state-of-the-art methods in terms of effectiveness.

\begin{figure*}
\centering
\includegraphics[width=0.7\textwidth]{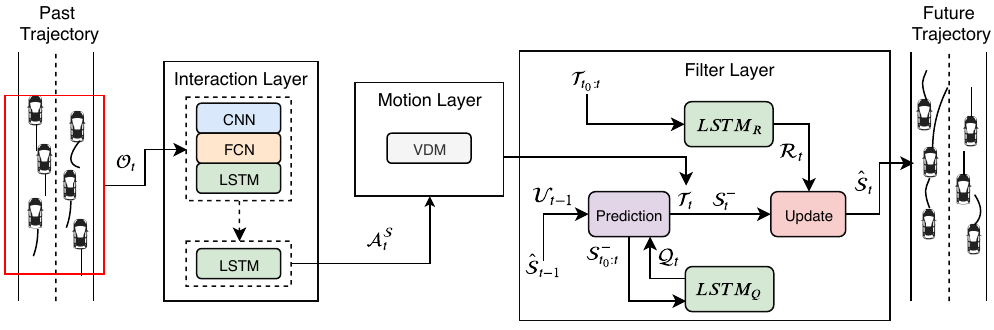}
\caption{Illustration of the \textbf{IaKNN} Model: In the diagram, at timestamp $t$, the environmental observation $\mathcal{O}_{t}$ flows into the interaction layer which generates the \emph{interaction-aware acceleration} $\mathcal{A}_{t}^{\mathcal{S}}$. Then, we calculate the \emph{interaction-aware trajectory} of vehicles $\mathcal{T}_t$ w.r.t Vehicle Dynamic Model (VDM) in motion layer. In the end, time-varying multi-agent Kalman neural networks run over the predicted time horizon $L$ to fuse dynamic trajectory $\mathcal{S}_t$ and \emph{interaction-aware trajectory} $\mathcal{T}_t$. Particularly, the time-varying process and measurement noises in the filter layer are set by zero-mean Gaussian noises with covariance formulated in a gated-structure neural network.}
\vspace{-2mm}
\label{figure1}
\end{figure*}

\section{Related Work}
\subsubsection{State Estimation}
State estimation is a mature subfield in robotics with the aim to estimate the state of a robot from various noisy measurements. One comprehensive survey of classic approaches of state estimation refers to \cite{barfoot2017state}. The limitation in traditional state estimation models is lack of \emph{prior} knowledge purified from database to be the preset parameter adapting time varying influences. Nowadays, neural network approaches have been explored largely for state estimation in autonomous industry. Coskun et al.~\cite{coskun2017long} train the triple-LSTM neural networks architecture to learn the kinematic motion model, process noise, and measurement noise for estimating human pose in a camera image. Haarnoja et al.~\cite{haarnoja2016backprop} adopt the discriminative approach in state estimation where neural networks is used to learn features from highly complex observations, and then filtered the features to underlying states. 

\subsubsection{Data-driven Approach Trajectory Prediction}
Trajectory prediction, which is a traditional topic in the field of intelligent vehicle society, has been largely studied, referring to the survey~\cite{lefevre2014survey,betts1998survey}. Among those methods for this topic, the data-driven ones are promising. For example, Ma et al.~ \cite{ma2018trafficpredict} propose an LSTM-based two-layers model TrafficPredict for heterogeneous traffic-agents in an urban environment. 

To increase robustness and accuracy in multi-agent tracking problems, the data-driven fashion can model more complex "interactions" between agents than the hand-crafted functions. For example, Alahi et al.~\cite{alahi2016social} propose a deep learning model to predict the motion dynamics of pedestrians in a crowded scene in which they build a fully connected layer called social pooling to learn the social tensor based on pedestrians. Gupta et al.~\cite{gupta2018social} propose a GAN-based encoder-decoder framework for trajectory prediction with a pooling mechanism to aggregate information across people. In the field of intelligent vehicles, Deo and Trivedi~\cite{deo2018convolutional} extract a social tensor with a convolutional social pooling layer and then feed the social tensor to a maneuver-based motion model for trajectory prediction. Kuefler et al.~\cite{kuefler2017imitating} and Bhattacharyya et al. \cite{bhattacharyya2018multi} use imitation learning approach to learn human drivers' behaviors for trajectory prediction. The learned policies are able to generate the future driving trajectories that match those of human drivers better and can also interact with neighboring vehicles in a more stable manner over long horizons. Zhao et al.~\cite{zhao2019multi} design an encoder-decoder architecture called \emph{multi-agent tensor fusion network} to extract multi-agent interactions with the spatial structure of agents and the scene context, and predict recurrently to agents' future trajectories.

Our IaKNN model differs from these models in two aspects. First, IaKNN captures the interactive effects in a form of accelerations which could then be feed to kinematics models and thus it enjoys the merits of both the classic Physics models and the data-driven process (of capturing the interactive effects). Second, IaKNN employs the Kalman filter for optimizing the state estimation, where LSTM neural networks are used for learning the time-varying process and measurement noises that are used in the Kalman model, and this is the first of its kind for trajectory prediction.

\section{Traffic Datasets}\label{trafficdataset}
To the best of our knowledge, there are four publicly available traffic datasets, namely Cityscapes~\cite{cordts2016cityscapes}, KITTI~\cite{geiger2013vision}, ApolloScape~\cite{huang2018apolloscape}, and NGSIM~\cite{colyar2007us}.
Cityscapes, KITTI and ApolloScape are collected from the first person perspective which have been widely adopted for single-agent systems in the field of intelligent vehicle society.
NGSIM, is collected on the southbound US101 road and the eastbound I-80 road with a software application called NG-VIDEO which transcribes vehicles' trajectories from an overhead video.
In this work, we only use NGSIM since among the 4 datasets, NGSIM is the only one that is suitable for a study in the multi-agent system which we target in this paper. Additionally, as reported in some existing studies~\cite{thiemann2008estimating,rahman2018real}, noises vary significantly in NGSIM, and this is one of the motivations that we proposed to learn the time-varying covariances in the Kalman filter.
\section{Kalman Filter}\label{Kalman}
In this part, we provide some background of the Kalman filter (KF) which is used as a building block in our model in this paper.
KF is an optimal state estimator in the mean square error (MSE) sense with a linear (dynamic) model and Gaussian noise assumptions. Suppose the state, control, and observation of the linear model are $s_t$, $u_t$ and $z_t$, respectively. The model could be expressed with a process equation and a measurement equation as follows.
\begin{align*}
s_t &= \mathcal{F} \cdot s_{t-1} + \mathcal{B} \cdot u_{t-1} + \omega, \\
z_t &= \mathcal{H} \cdot s_t + \eta,
\end{align*}
where $\mathcal{F}$ is a dynamic matrix, $\mathcal{B}$ is a control matrix, $\mathcal{H}$ is an observation matrix, which are all known.
Moreover, $\omega \sim \mathcal{N}(0, \mathcal{Q})$ is the process noise
and $\eta \sim \mathcal{N}(0,\mathcal{R})$ is the measurement noise
based on the noise covariance matrices $\mathcal{Q}$ and $\mathcal{R}$, respectively.

The process of KF is as follows.
It iterates between a prediction phase and an update phase for each of the observations. In the prediction phase, the current state $s^-_t$ and the error covariance matrix $\mathcal{P}^-_t$ are estimated as follows.
\begin{align*}
s_t^- &=  \mathcal{F} \cdot \hat{s}_{t-1} + \mathcal{B} \cdot u_{t-1},\\
\mathcal{P}^-_t &= \mathcal{F} \cdot \hat{\mathcal{P}}_{t-1}\cdot \mathcal{F}^T + \mathcal{Q}.
\end{align*}
In the update phase, once the current observation $z_t$ is received,
the Kalman gain $\mathcal{K}_t$, the prior estimation $\hat{s}_t$ and the error covariance matrix $\hat{\mathcal{P}}_t$ are calculated as follows.
\begin{align*}
\mathcal{K}_t &= \mathcal{P}^-_t \cdot \mathcal{H}^T \cdot (\mathcal{H} \cdot \mathcal{P}^-_t \cdot \mathcal{H}^T + \mathcal{R})^{-1},\\
\hat{s}_t & = s^-_t  + \mathcal{K}_t \cdot(z_t - \mathcal{H} \cdot s^-_t),\\
\hat{\mathcal{P}}_t &= ( I - \mathcal{K}_t \cdot \mathcal{H}) \cdot \mathcal{P}^-_t.
\end{align*}
For a comprehensive review of KF, the readers could refer to standard references \cite{bishop2001introduction}.

KF is effective and commonly used as a basic data processing skill in autonomous industry where sophisticated KFs, for instance, Extended KF (EKF) and Particle Filtering (PF) are also adopted in the literature~\cite{crassidis2007survey,aulinas2008slam}. In our work, we take a novel approach only adopting KF as the filtering part in neural-network based filtering algorithm. It is worth mention that, the filter we used is in some sense an advanced version of KF where the noise covariances are being learned online but not pre-set.

\section{Problem Statement}\label{problemstatement}

We assume there are $N$ vehicles in the multi-agent system (traffic scene). For each vehicle at timestamp $t$, we collect its position $p_t$, velocity $v_t$, acceleration $a_t$, vehicle width $w_t$, vehicle length $l_t$, and relative distances $\{d_t^j\}_{j=1}^{N-1}$ with other agents. We call the observations of all vehicles \emph{environmental observation at timestamp $t$} denoted as $o_t$. Given the past $h$-length environmental observations $\mathcal{O}_t := \{o_{t-h+1}, o_{t-h+2}, \cdots, o_{t}\}$, we aim to predict the future $L$-length trajectories of each vehicle.

Note that the number of vehicles $N$ is set to be 6 in our experiments because our setting is the two-lane dynamics where the ego-vehicle has one in front, one at the rear, and three vehicles in the neighboring lane. $N$ is an independent variable in our methodology, and thus it is straightforward to increase $N$ to consider more vehicles for trajectory prediction tasks.

\section{Methodology}\label{Method}
In this section, we present our architecture \emph{interaction-aware Kalman neural networks} (IaKNN). Figure~\ref{figure1} gives an overview of the architecture, where the notations are explained as follows. $\mathcal{A}^{\mathcal{S}}$ is the portfolio of interaction-aware accelerations outputted by the interaction layer. $\mathcal{T}$ is the portfolio of interaction-aware trajectories computed by the motion layer, and $\mathcal{S}$ and $\mathcal{V}$ are the state and the control of the Kalman filter in the filter layer, respectively, where $\mathcal{R}$ and $\mathcal{Q}$ are the noise covariance matrices, both learned by LSTM neural networks. Besides, in this paper, $t_0$, $t$, $L'$ and $L$ represent the starting time, current time, observation time horizon and prediction time horizon, respectively, where $t_0 \leq t \leq t_0+L'$.
%

In the following, we present three layers of IaKNN, namely the \textbf{interaction layer}, the \textbf{motion layer} and the \textbf{filter layer}. The notations that are frequently used throughout the paper are given in Table \ref{notation}.

\begin{table}
  \centering
  \small{
  \setlength{\belowcaptionskip}{1pt}
  \caption {Notations and meanings (at timestamp $t$).}\label{notation}
  \begin{tabular}{c|c}
     \hline
          {\bf Notation}     & {\bf Meaning}\\
     \hline\hline
          $\mathcal{O}_t$  & Traffic Environment Observation\\
     \hline
          $\mathcal{A}_t^{\mathcal{S}}$  & Interaction-aware Acceleration\\
     \hline
          $\mathcal{T}_t$  &  Interaction-aware Trajectory \\
     \hline
          $\mathcal{F}$ & State Transition Matrix\\
     \hline
          $\mathcal{B}$ & Control Matrix\\
     \hline
          $\mathcal{S}_t$ & Dynamic Trajectory\\
     \hline
         $\mathcal{S}^-_t$ &\emph{Priori} Estimation of Dynamic Trajectory\\
      \hline
         $\hat{\mathcal{S}}_t$ & \emph{Posteriori} Estimation of Dynamic Trajectory\\
     \hline
          $\mathcal{U}_t$ & Dynamic Acceleration\\
     \hline
          $\mathcal{Q}_t$ & Process Noise Covariance\\
     \hline
          $\mathcal{R}_t$ & Measurement Noise Covariance\\
     \hline
          $G_t$ & Ground Truth of Future Trajectory\\
     \hline
          $N$ & Number of Vehicles\\
     \hline
         $L/L'$ & Observation/Prediction Time Horizon\\
     \hline
  \end{tabular}}
\end{table}



\subsection{Interaction Layer}\label{Interaction}
In the interaction layer, we aim to extract the \emph{interaction-aware accelerations} $\mathcal{A}^{\mathcal{S}}$ from the past traffic environment observations $\mathcal{O}_t$.
\subsubsection{Interaction-aware Acceleration}

Normally, the motion of a vehicle would be determined by its own vehicle dynamics.
Nevertheless, in a multi-agent system which we target in this paper,
the situation is much more complex
since drivers of vehicles would be affected by those of other vehicles that are nearby (or they would interact with one another).
For example, a vehicle would be forced to slow down if another vehicle nearby tries to cut the lane in the front.
In fact, the motion of vehicles is determined by not only their physical accelerations but also the interactive effects among vehicles.
Inspired by the classical social force model \cite{helbing2000simulating},
which models the intention of a driver to avoid colliding with dynamic or static obstacles,
we propose to extract those accelerations
such they capture both the raw accelerations recorded and the interactions among vehicles nearby.
We call them the \emph{interaction-aware accelerations} and denote them by $\mathcal{A}^{\mathcal{S}}$.

\begin{figure}
	\centering
	\includegraphics[width=0.5\textwidth]{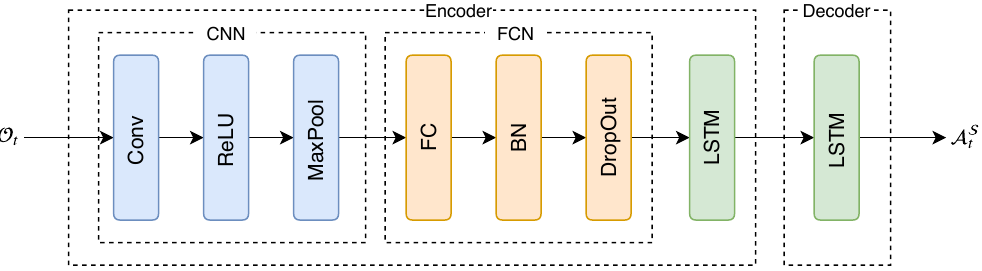}
	\caption{Illustration of \textbf{Interaction Layer}}
	\vspace{-2mm}
	\label{figure2}
\end{figure}

Specifically, at timestamp $t$, traffic environment observations $\mathcal{O}_{t}$ includes a sequence of recorded accelerations $a_{t_0:t}$, vehicle widths $w_{t_0:t}$, vehicle lengths $l_{t_0:t}$, and relative distances $d_{t_0:t}$ of agents in the system.
By following \cite{helbing1995social},
we compute the so-called \emph{repulsive interaction forces} $e_{t_0:t} := \text{exp}\big((v^i +v^j)\cdot \Delta t - d^{ij}\big)_{t_0:t}$, where superscripts $i$ and $j$ represent two vehicles that are close to each other and
include them in $\mathcal{O}_{t}$.
Thus, the interaction operator formula at timestamp $t$ is written in details as,
\begin{align*}
\mathcal{A}_{t}^{\mathcal{S}} =  \textbf{Interaction}_{\{\mathcal{W}, b\}}\big(a_{t_0:t}, w_{t_0:t}, l_{t_0:t}, d_{t_0:t}, e_{t_0:t}\big).
\end{align*}
The interaction layer is implemented as a neural network as presented in Figure \ref{figure2}. The architecture of interaction layer is an LSTM encoder-decoder. In the encoder, we build convolutional layers (CNN) regarded as a social tensor extractor,  fully-connected layers (FCN) regarded as a mixer of the social features, and merge the deep features into the encoder LSTM. In the decoder, the decoder LSTM outputs the predicted accelerations. Note that we applied the batch normalization to LSTMs with the vertical connections, which are transported from one layer to another, and it is technologically similar to some existing studies~\cite{laurent2016batch,amodei2016deep}. In addition, we introduce the DropOut (with the fraction of 0.5) in case of overfitting.

\subsubsection{Operator Representation}
At timestamp $t$, the interaction layer in an operator formula is written as,
\begin{align*}
     \textbf{Interaction}_{\{\mathcal{W}, b\}}: \mathcal{O}_{t} \longmapsto \mathcal{A}_{t}^{\mathcal{S}},
\end{align*}
where $\mathcal{O}_{t}$ is a portfolio of past environmental observations from $t_0$ to $t$ and $\mathcal{A}_{t}^{\mathcal{S}}$ is the portfolio of the interation-aware accelerations from $t+1$ to $t+L$.

\subsection{Motion Layer}\label{Motion}
In the motion layer, we aim to calculate the \emph{interaction-aware trajectories} $\mathcal{T}$ based on the \emph{interaction-aware accelerations} $\mathcal{A}^\mathcal{S}$ from the interaction layer.

The main intuition of the motion layer comes from the primary kinematic equation which establishes a relationship among \textit{position, time and velocity}. Our strategy is to use higher-order derivatives of a position for better forecasting. Specifically, let $p_t$ be the position of a dynamic obstacle at timestamp $t$. We write $p_t$ with the Taylor expansion as follows.
\begin{align}
p_t = p_{t-1} + v_{t-1}\cdot \Delta t + \frac{1}{2} a_{t-1}\cdot \Delta t^2 + O\Big(\Delta t^3\Big) \label{taylor}
\end{align}
where $v_{t-1}$ represents the velocity at timestamp $t-1$,
$a_{t-1}$ represents the acceleration at timestamp $t-1$, and
the Big-O term captures all remaining terms which would be ignored.
Moreover, we replace the acceleration term $a_t$ with an {\it interaction-aware acceleration} $\mathcal {A^S}$
which is derived from the environment observations.
%

We specify the velocity term $v$ in Equation~\ref{taylor} as follows.
Suppose the current timestamp is $t$.
For $v_t$, we take the velocity readings which are currently available and transform them to $v_t$ by using a dynamic model - depending on the agent type, we adopt different dynamic models for this task, which shall be introduced shortly.
For $v_{t+1}, v_{t+2}, ...$,
we estimate their values by applying an integral function based on the interaction-aware accelerations as follows.
\[
v_{t+i}:= \int_t^{t+i}{\mathcal{A}^{\mathcal{S}}}dt, 
\]
where $i = 1, 2, ..., L$.

Next, we introduce the dynamic models, vehicle dynamic model (VDM), which map motion along the axes of the global reference frame to motion along the axe fo the robot's local reference frame. By following \cite{pepy2006path},
we implement the vehicle dynamic model as a classical bicycle model \cite{taheri1992investigation}.
Specifically, suppose $s := (x, y, \theta, v_x, v_y, r)$ is the current reading involving velocities, where $x$ and $y$ are the coordinates, $\theta$ is the orientation, $v_x$ and $v_y$ are velocities,
and $r$ is the yaw rate.
The bicycle model is written as follows.
\begin{align*}
\dot{x} &= v_x\cdot \cos{\theta} -v_y \cdot \sin{\theta},\\
\dot{y} &= v_x\cdot \sin{\theta} + v_y \cdot \cos{\theta},\\
\dot{\theta} &= r.
\end{align*}
$\dot{x}$ and $\dot{y}$ are the transformed velocities along the $x$ and $y$ directions, respectively.
For more details, the readers are referred to \cite{pepy2006path}. Note that the nonlinear VDM model is not being estimated in the Kalman filter, and it is used only to transform the velocity readings which are based on a vehicle-centric coordinating system to those based on a global coordinating system. That is, VDM model is simply a plug-in transformation function, and Kalman filter used in this paper is based on a linear system (on positions and velocities).

To simplify this layer, we regard all agents as mass points and their motion behaviors are described in the basic kinematic motion equations $\dot{x} = v_x$ and $\dot{y} = v_y$.

\subsubsection{Operator Representation}
At timestamp $t$, the motion layer in an operator formula is written as,
\begin{align*}
     \textbf{Motion}: \mathcal{A}_{t}^{\mathcal{S}}  \longmapsto \mathcal{T}_{t},
\end{align*}
where $\mathcal{T}_{t}$ is an interaction-aware trajectory from $t+1$ to $t+L$.

\subsection{Filter Layer}\label{Filter}
In the filter layer, we establish a model
based on the Kalman filter
to estimate the dynamic trajectories $\mathcal{S}_{t}$ based on the \emph{interaction-aware trajectories} $\mathcal{T}_{t}$ used as observations.

\subsubsection{Filter Model}
To fit the Kalman filter as described in Section \ref{Kalman},
we let the dynamic trajectories $\mathcal{S}_t$ be the states, the \emph{interaction-aware trajectories} $\mathcal{T}_t$ be the observations and the dynamic accelerations $\mathcal{U}_{t}$ be the controls in a linear model.
As a result,
the equation of the linear dynamic model could be written as follows.
\begin{align}
\mathcal{S}_t &= \mathcal{F}\cdot \mathcal{S}_{t-1} + \mathcal{B}\cdot \mathcal{U}_{t-1} + \omega_t, \label{kf1}\\
\mathcal{T}_t &= \mathcal{S}_{t} + \eta_t \label{kf2}
\end{align}
where $\mathcal{F}$ is the state transition matrix,
$\mathcal{B}$ is the control matrix,
$\omega_t \sim \mathcal{N}\big(0, \mathcal{Q}_t\big)$ are the time-varying process noises and
$\eta_t \sim \mathcal{N}\big(0,\mathcal{R}_t\big)$ are the measurement noises.
The time-varying covariances $\mathcal{Q}_t$ and $\mathcal{R}_t$ will be learned by time-varying noise models
which consist of LSTM neural networks and will be introduced later.
Note that here we assume the observation matrix $\mathcal{H}$ is an identity matrix for simplicity.


\subsubsection{Specifications of the Layer}
We assume $N$ agents (dynamic obstacles) in the multi-agent system. At timestamp $t$, the state $\mathcal{S}_t$ and the observation $\mathcal{T}_t$ of our Equation \ref{kf1} and \ref{kf2} could be written as follows.
\[
\mathcal{S}_t :=
\begin{bmatrix}
\mathcal{S}_t^1\\
\vdots \\
\mathcal{S}_t^N\\
\end{bmatrix}_{(2\cdot N \cdot L )\times1} \text{and\hspace*{1em}}
\mathcal{T}_t:=
\begin{bmatrix}
\mathcal{T}_t^1\\
\vdots \\
\mathcal{T}_t^N  \\
\end{bmatrix}_{(2\cdot N \cdot L )\times1},
\]
where the state $\mathcal{S}_t^i$ includes positions $p_k^i$ from GPS and velocities $v_k^i$ from the wheel odometer and the observation $\mathcal{T}_t^i$ includes the predicted positions $\bar{p}_k^i$ and the predicted velocities $\bar{v}_k^i$, where $t + 1 \leq k \leq t + L$.
Specifically, we have the following.
\[
\mathcal{S}_t^i :=
\begin{bmatrix}
p_{t+1}^i\\
v_{t+1}^i\\
\vdots\\
p_{t+L}^i\\
v_{t+L}^i\\
\end{bmatrix}_{(2\cdot L)\times1}  \text{and\hspace*{1em}}
\mathcal{T}_t^i :=
\begin{bmatrix}
\bar{p}_{t+1}^i\\
\bar{v}_{t+1}^i\\
\vdots\\
\bar{p}_{t+L}^i\\
\bar{v}_{t+L}^i\\
\end{bmatrix}_{(2\cdot L)\times1},
\]
where the subscript $L$ is the predicted time horizon.

Next, we define the state transition matrix $\mathcal{F}$ and the control matrix $\mathcal{B}$ in our model.
Firstly, we define two matrix blocks $M_1$ and $M_2$ as follows.
\[
M_1 :=
\begin{bmatrix}
1 & \Delta t &  &  & \\
 &  1 &  &  & \\
 &  &  \ddots &  & \\
 &  &  &  1 & \Delta t\\
 &  &  &  & 1
\end{bmatrix}_{(2\cdot L)\times (2 \cdot L)}
\]
and
\[
M_2 :=
\begin{bmatrix}
 \frac{1}{2}\Delta t^2 &  &  \\
 \Delta t &  &  \\
 &   \ddots  & \\
 &    & \frac{1}{2}\Delta t^2 \\
 &    & \Delta t
\end{bmatrix}_{(2\cdot L)\times L}
\]
where $\Delta t$ is the time difference between two adjacent traffic environment observations.
Then, $\mathcal{F}$ and $\mathcal{B}$ are block diagonal matrices that are defined as follows.
\[
\mathcal{F} := diag\big(\underbrace{M_1, \dots,M_1}_{N}\big), \text{ and }\mathcal{B} := diag\big(\underbrace{M_2, \dots,M_2}_{N}\big).
\]
\subsubsection{Prediction and Updated Steps}
The prediction step of the Kalman filter is defined as,
\begin{align*}
\mathcal{S}_t^- &= \mathcal{F}\cdot \hat{\mathcal{S}}_{t-1} + \mathcal{B} \cdot \mathcal{U}_{t-1},\\
\mathcal{P}^-_t &= \mathcal{F}\cdot \hat{\mathcal{P}}_{t-1}\cdot \mathcal{F}^T + \mathcal{Q}_t,
\end{align*}
and the update step is as,
\begin{align*}
\mathcal{K}_t &= \mathcal{P}^-_t\cdot (\mathcal{P}^-_t +\mathcal{R}_t)^{-1},\\
\hat{\mathcal{S}}_t & = \mathcal{S}^-_t  + \mathcal{K}_t\cdot(\mathcal{T}_t - \mathcal{S}^-_t),\\
\hat{\mathcal{P}}_t &= (\mathcal{I}-\mathcal{K}_t) \cdot \mathcal{P}^-_t,
\end{align*}
where $\mathcal{Q}_t$ and $\mathcal{R}_t$ are the outputs of the time-varying noise models that we introduce next.
\subsubsection{Time-varying Noise Model}\label{tvn}
Since our desired filter model is time-varying,
we assume both the process noises and the measurement noises to follow a zero-mean Gaussian noise model with its covariances formulated as $\mathcal{Q}_t:= LSTM_Q\big(\mathcal{S}_{t_0:t}^-\big)$ and $\mathcal{R}_t:=LSTM_R\big(\mathcal{T}_{t_0:t}\big)$, respectively. Here, we adopted LSTM because the noises are generated sequentially with trajectories and LSTM is known for its capability of capturing sequential dynamics \cite{wang2019effective}. The concrete $\mathcal{Q}_t$ and $\mathcal{R}_t$ are tractable via backpropagation under the end-to-end trainable architecture. Besides, $\mathcal{Q}_t$ and $\mathcal{R}_t$ are not PSD in general, but in practice, it is possible to implement a LQ decomposition to guarantee the PSD property of $\mathcal{Q}_t$ and $\mathcal{R}_t$ as some existing studies do \cite{haarnoja2016backprop}.

\subsubsection{Operator Representation}
At timestamp $t$, the filter layer in an operator formula is written as,
\begin{align*}
    \textbf{Filter}_{\{\mathcal{W}, b\}}: \big\{\hat{\mathcal{S}}_{t-1},  \mathcal{T}_{t}, \mathcal{U}_{t-1} \big\} \longmapsto \hat{\mathcal{S}}_{t},
\end{align*}
where $\hat{\mathcal{S}}_{t}$ is the \emph{Posteriori} estimation of dynamic trajectory from $t+1$ to $t+L$.
\subsection{Loss Function}
The loss function of IaKNN (\textbf{Interaction}$_{\{\mathcal{W}, b\}}$, \textbf{Motion}, and \textbf{Filter}$_{\{\mathcal{W}, b\}}$) is defined as the sum of displacement error of \emph{Posteriori} estimation $\hat{\mathcal{S}}$ of dynamic trajectories and ground truth $G$ over all time steps and agents, as follows.
\[
\mathcal{L}_{\{\mathcal{W}, b\}}:= \frac{1}{(L'+1)\cdot N} \cdot \sum_{i=1}^N \sum_{t=t_0}^{t_0+L'} ||\hat{\mathcal{S}}_t^i  - G^i_t||^2,
\]
where $G^i_t$ means the ground truth of the future trajectory of $i$-th agent at the start timestamp $t$. Noth that $L'$ is the observation time horizon as defined in the above.


\begin{figure}
\begin{tabular}{c c}
\begin{minipage}{4cm}
\centering
\includegraphics[width=4cm]{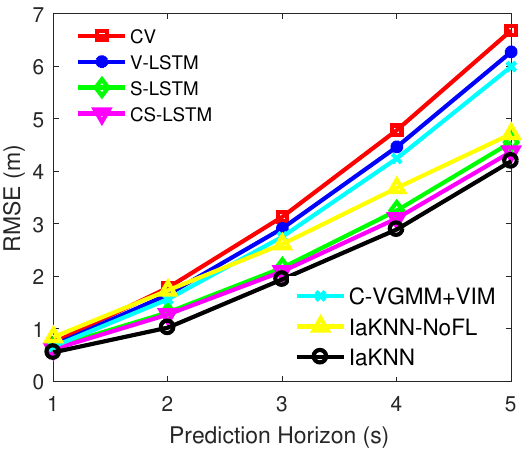}
\end{minipage}
&
\begin{minipage}{4cm}
\centering
\includegraphics[width=4cm]{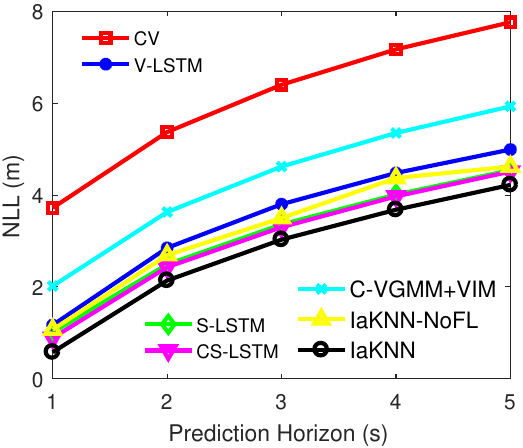}
\end{minipage}
\end{tabular}
\caption{Illustration of RMSE and NLL of model \emph{CV, V-LSTM, S-LSTM, C-VGMM+VIM, CS-LSTM, IaKNN-NoFL, and IaKNN}. For both evaluation metrics, we plot its average for the prediction time horizon in 5s.}
\vspace{-6mm}
\label{effect}
\end{figure}

\section{Experiments}
\subsection{Dataset} \label{dataset}
We evaluate our approach IaKNN on two public datasets,
namely US Highway 101 (US-101) and Interstate 80 (I-80) of the NGSIM program.
Each dataset contains $(x, y)$-coordinates of vehicle trajectories in a real highway traffic with 10Hz sampling frequency over a 45-min time span.
The 45-min dataset consists of three 15-min segments of mild, moderate and congested traffic conditions.
We follow the experimental settings that were proposed by existing studies \cite{deo2018multi,deo2018convolutional} and
combine US-101 and I-80 into one dataset.
As a result, the dataset involves 100,000 frames of raw data.

We construct the multi-agent training traffic scene in the following construction procedure.
Firstly, we align the raw data by its timestamps.
Secondly, we form a multi-agent traffic scene by picking one host vehicle and including five closest vehicles on its traffic lane or two adjacent traffic lanes.
Finally, we set the window size for extraction as 7 seconds to generate the training scenes. We will explain the setting of window size in the next section.


\subsection{Baselines}
The following baseline models will be compared with our model IaKNN.
\begin{itemize}[leftmargin=10pt]
\item \emph{Constant Velocity} (\textit{CV}): Model of the primary kinematic equation with constant velocity.
\item \emph{Vanilla-LSTM} (\textit{V-LSTM}): Model of Seq2Seq. It is from a sequence of past trajectories to a sequence of future trajectories~\cite{park2018sequence}.
\item \emph{Social LSTM} (\textit{S-LSTM}): Model of LSTM-based neural network with a \textit{social pooling} for pedestrian trajectory prediction. As demonstrated in~\cite{alahi2016social}, the model performs consistently better than traditional models such as the linear model, the collision avoidance model and the social force model. Therefore, we do not compare these traditional methods in our experiments.
\item \emph{C-VGMM+VIM}: One variational Gaussian mixture models with Markov random fields, taken into account the interaction effects between agents~\cite{deo2018would}. 
\item \emph{Convolutional Social Pooling-LSTM} (\textit{CS-LSTM}): Maneuver based motion model which will generate a multi-modal predictive distribution~\cite{deo2018convolutional}.
\item \emph{IaKNN-NoFL}: The proposed method IaKNN without the filter layer.
\end{itemize}
Note that the model \emph{GAIL-GRU} is not taken into account in the comparison, because it has access to the future ground-truth trajectories of neighboring vehicles to predict the ego-vehicle's trajectory, while the others have not~\cite{kuefler2017imitating}.

\subsection{Evaluation Metrics}
Two metrics, namely the \textit{root-mean-square error} (RMSE) and \textit{negative log-likelihood} (NLL), are used to measure the effectiveness of IaKNN.
In particular, the first 2-seconds trajectories and the rest 5-seconds trajectories are used as past trajectories and the ground truth in a 7-seconds~\footnote{The window size is equal to the sum of the prediction horizon and the observation. We set the prediction horizon to be 5 seconds by following the work of CS-LSTM~\cite{deo2018convolutional}. Regarding the observation horizon, existing algorithms used different settings as follows: Social LSTM (3.2 seconds), CS-LSTM (3 seconds), Trafficpredict (2 seconds). We set the observation horizon to be 2 seconds, resulting in a window size of 7 seconds.} multi-agent training traffic scene, respectively.
\begin{itemize}[leftmargin=10pt]
\item RMSE: the root mean squared sum accumulated by the displacement errors over the predicted positions and real positions during the prediction time horizon.
\item NLL: the sum of the negative log probabilities of the predicated positions against the ground-truth positions during the prediction time horizon (we consider a predicted position to be correct if its distance from the ground-truth one is bounded by a small threshold and wrong otherwise).
\end{itemize}

\subsection{Implementation Details}
The default length of the past trajectories is two seconds and
the time horizon of the predicted trajectories is one to five seconds.
The default number of hidden units in LSTMs in the interaction layer and filter layer is set to 32 and all LSTM weight matrices are initialized using a uniform distribution over $[-0.001, 0.001]$.
The weight matrices for other layers are set with the Xavier initialization.
The biases are initialized to zeros.
Additionally, in the training process, we adopt the Adam stochastic gradient descent with hyper-parameters $\beta_1 = 0.9$, $\beta_2 = 0.99$ and set the initial learning rate to be 0.001.
In order to avoid the gradient vanishing, a maximum gradient norm constraint is set to 5.
For the parameters of baselines, we follow the original settings in the open source code.
The experiments are conducted on a machine with Intel(R) Xeon(R) CPU E5-1620 and one NVIDIA GeForce GTX 1070 GPU.

\subsection{Result Analysis}
The performance results of baselines and our method on the traffic scene are shown in Figure \ref{effect}. We compute the RMSE and NLL for all traffic scenes and plot the average for the prediction time horizon in 5s. The naive \textit{CV} produces the largest prediction errors in all comparison methods. \textit{V-LSTM, S-LSTM} and \textit{CS-LSTM} perform similarly in terms of both metrics which is mainly because they are all LSTM-based neural networks. Additionally, \textit{S-LSTM, CS-LSTM}, and \textit{IaKNN-NoFL} perform better than \textit{V-LSTM} and \textit{C-VGMM+VIM}, especially in RMSE, and this is mainly because the formulation establishes on multi-agent system setting, which also takes into account the interactive effects among vehicles in modeling. IaKNN outperforms slightly all baselines in terms of both metrics. Specifically, we observe that it outperforms the best baseline \textit{CS-LSTM} with about 10\% improvement. This may be explained by the fact that the filter layer in our IaKNN model estimates the underlying trajectories from both interaction-aware trajectories $\mathcal{T}$ and dynamic trajectories $\mathcal{S}$ in a traffic scene and the interaction layer has done a good job in capturing the interactive effects among the surrounding vehicles. The combination of the deep neural network and probabilistic filter makes our model more applicable for the real-time trajectory prediction in the traffic scene.

\subsection{Case Studies}
\begin{figure}[htpb]
\begin{tabular}{c}
\begin{minipage}{8cm}
\centering
\includegraphics[width=6.5cm]{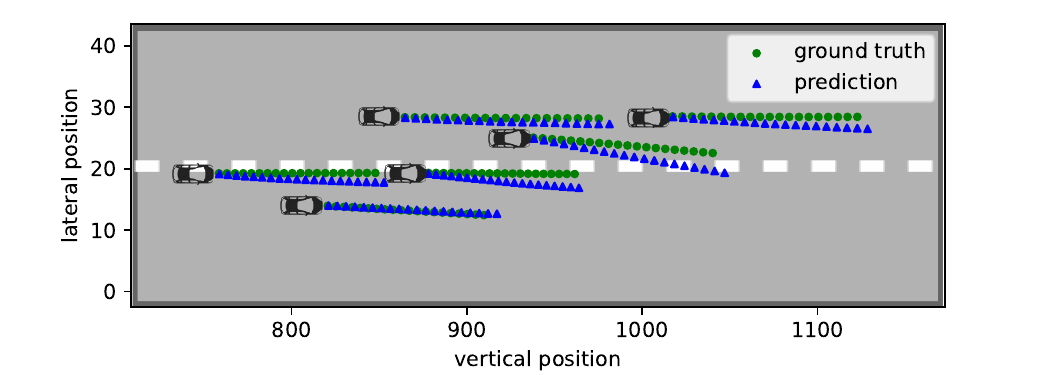}
\end{minipage}
\\
\begin{minipage}{8cm}
\centering
\includegraphics[width=6.5cm]{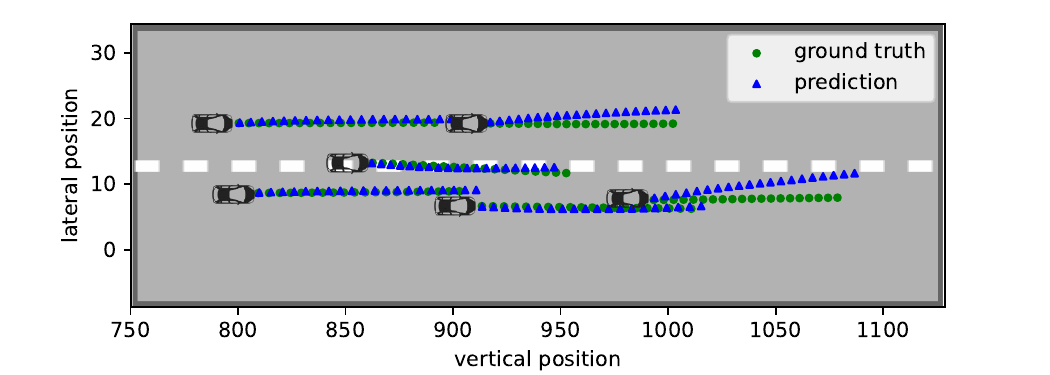}
\end{minipage}
\end{tabular}
\caption{Case studies of the prediction result by IaKNN: The predicted trajectories and the real ones are drawn in blue and green color, respectively. For each vehicle, we plot its future 2s trajectory.}
\vspace{-1mm}
\label{case}
\end{figure}

We illustrate the results by showing the predicted 2-second trajectories of vehicles by IaKNN and the real ones in the two lane-change traffic scenarios in Figure \ref{case}. The results demonstrate the effectiveness of the prediction by IaKNN intuitively. We observe that the predicted trajectories (color blue) are close to the real ones (color green) for all vehicles on the lanes in the Figure \ref{case}.

To prevent the traffic accidents, vehicles keep a safe following distance from others. The predicted trajectories by IaKNN as shown Figure \ref{case} confirm this statement, which is one of the clear superiorities that the multi-agent model outperforms the single-agent model in the traffic trajectory prediction scene. Specifically, we observe predicted trajectories of the front vehicles by IaKNN in Figure \ref{case} have clear intentions to change their lanes, one way to increase the safety following distance, in the future seconds. Changing lane is one of the crucial issues that an autonomous vehicle need to take actions to manage the traffic risks before it happens. Hence, IaKNN is a more sensitive traffic prediction algorithm than the past algorithms in autonomous vehicle industry based on the comprehensive modeling in an interaction-aware multi-agent environment. 

\section{Conclusion and Future Work}
In this study, we propose an architecture, IaKNN, to predict the motion of surrounding vehicles in a dynamic environment, in which we make the first attempt to generate an intractable quantity from complex traffic scene yielding a new interaction-aware motion model. Extensive experiments show that IaKNN outperforms the baseline models in terms of effectiveness on the NGSIM dataset. Further work will be carried out to extend IaKNN to a probabilistic formulation and combine IaKNN with a maneuver-based model in which road topology and the traffic information are taken into account.

\section{Aknowledgement}
We would like to thank WeBank FATE developer community, Baidu Apollo developer community, Yuan Jin (Shenzhen Gradient Technology Co., Ltd.) and Ruihui Zhao (Tencent Jarvis Lab) for their useful suggestions and contributions.

{\footnotesize
\bibliographystyle{unsrt}
\bibliography{refs}
}

\end{document}